\documentclass{article} 
\usepackage{collas2023_conference,times}
\usepackage{easyReview}

\usepackage{amsmath,amsfonts,bm}

\def\eqref#1{equation~\ref{#1}}

\def\1{\bm{1}}

\DeclareMathAlphabet{\mathsfit}{\encodingdefault}{\sfdefault}{m}{sl}
\SetMathAlphabet{\mathsfit}{bold}{\encodingdefault}{\sfdefault}{bx}{n}

\usepackage{hyperref}
\usepackage{wrapfig}
\hypersetup{
    colorlinks=true,
    linkcolor=red,
    filecolor=magenta,
    urlcolor=blue,
    citecolor=purple,
    pdftitle={Overleaf Example},
    pdfpagemode=FullScreen,
    }

\title{Loss of Plasticity in Continual\\ Deep Reinforcement Learning}

\author{Zaheer Abbas\thanks{Corresponding author: \texttt{zaheersm@google.com}.} \ $^1$, Rosie Zhao$^2$, Joseph Modayil$^1$, Adam White$^{1,3,4,5}$, Marlos C. Machado$^{3,4,5}$\\
$^{1}$ DeepMind \ \
$^{2}$ Harvard University \ \
$^{3}$ University of Alberta\\
$^{4}$ Alberta Machine Intelligence Institute (Amii) \ \
$^{5}$ CIFAR AI Chair
}

\def\visit{visit}
\def\visits{visits}

\def\visiting{visiting}
\def\revisits{revisits}
\def\revisiting{revisiting}

\preprintcopy 

\begin{document}

\maketitle

\begin{abstract}
The ability to learn continually is essential in a complex and changing world.
In this paper, we characterize the behavior of canonical value-based deep reinforcement learning (RL) approaches under varying degrees of non-stationarity.
In particular, we demonstrate that deep RL agents lose their ability to learn good policies when they cycle through a sequence of Atari 2600 games.
This phenomenon is alluded to in prior work under various guises---e.g., loss of plasticity, implicit under-parameterization, primacy bias, and capacity loss.
We investigate this phenomenon closely at scale and analyze how the weights, gradients, and activations change over time in several experiments with varying dimensions (e.g., similarity between games, number of games, number of frames per game), with some experiments spanning 50 days and 2 billion environment interactions.
Our analysis shows that the activation footprint of the network becomes sparser, contributing to the diminishing gradients.
We investigate a remarkably simple mitigation strategy---Concatenated ReLUs (CReLUs) activation function---and demonstrate its effectiveness in facilitating continual learning in a changing environment.\looseness=-1

\end{abstract}

\section{Introduction}

Reinforcement learning (RL) approaches have been successfully used in several real-world applications, from discovering superior algorithms for video compression ~\citep{Mandhane22MuZero} and matrix multiplication ~\citep{Fawzi22Discovering} to designing superior floorplans for Google’s tensor processing units ~\citep{Mirhoseini21Graph}; from efficiently controlling cooling systems ~\citep{Luo22Controlling} and thermal power generators ~\citep{Zhan22DeepThermal} and managing inventories ~\citep{Madeka22Deep} to navigating balloons in the stratosphere~\citep{bellemare2020autonomous} and automatically producing plasma configurations for Tokamak nuclear fusion reactors ~\citep{Degrave22Magnetic}. 

Largely, these systems can be characterized as a search for a fixed policy that once deployed no longer changes.
The search is typically done offline using a batch of historical data or more often on a simulator \citep[e.g.,][]{Madeka22Deep, Zhan22DeepThermal, Mandhane22MuZero, Fawzi22Discovering, Degrave22Magnetic, bellemare2020autonomous}.
Indeed, in some cases the pursuit of a stationary policy is wholly appropriate as the policy does not need to change once deployed (e.g., a policy that multiplies matrices with the desired computational complexity).
But there are many applications in which a stationary policy can quickly become ineffective once the system is deployed and more data become available.
In cooling-system controllers, for instance, change in weather patterns and sensor drift requires continual learning to produce effective policies ~\citep{Luo22Controlling}.
More generally, continual adaptation is unavoidable in situations in which the world around the learning system continually changes~\citep{Sutton07Tracking}.

Today, most notable applications of reinforcement learning in the real-world use neural network function approximation (i.e., deep reinforcement learning).
Much of the neural network toolbox for function approximation was perfected for stationary settings with fixed data sets (i.e., supervised learning) that allow for independent and identically distributed (IID) sampling~\citep{hadsell2020embracing}.
Remarkably, recent work has shown that neural networks learn poorly when data become available incrementally~\citep{Ash20Warm}. More critically, when confronted with distribution shifts over time, neural networks lose their ability to learn altogether---a phenomenon characterized as \textit{loss of plasticity}~\citep{Dohare21Continual}.  

Perhaps unsurprisingly, when neural networks were first applied successfully to play Atari 2600 games with reinforcement learning---an inherently non-stationary problem as the data distribution shifts with changing policies---tools such as large replay buffers were employed to help approximate the stationary regime~\citep{Mnih2015Human}.
But long-lived reinforcement learning agents that are embedded in a complex world must deal with non-stationarities that manifest over timescales much longer than what can feasibly be stored in a replay buffer~\citep{Lesort20Continual, Luo22Controlling}, emphasizing the need for continual learning solutions that can work across timescales. 

\begin{wrapfigure}{r}{0.42\textwidth}
\begin{center}
\centerline{\includegraphics[width=0.35\columnwidth]{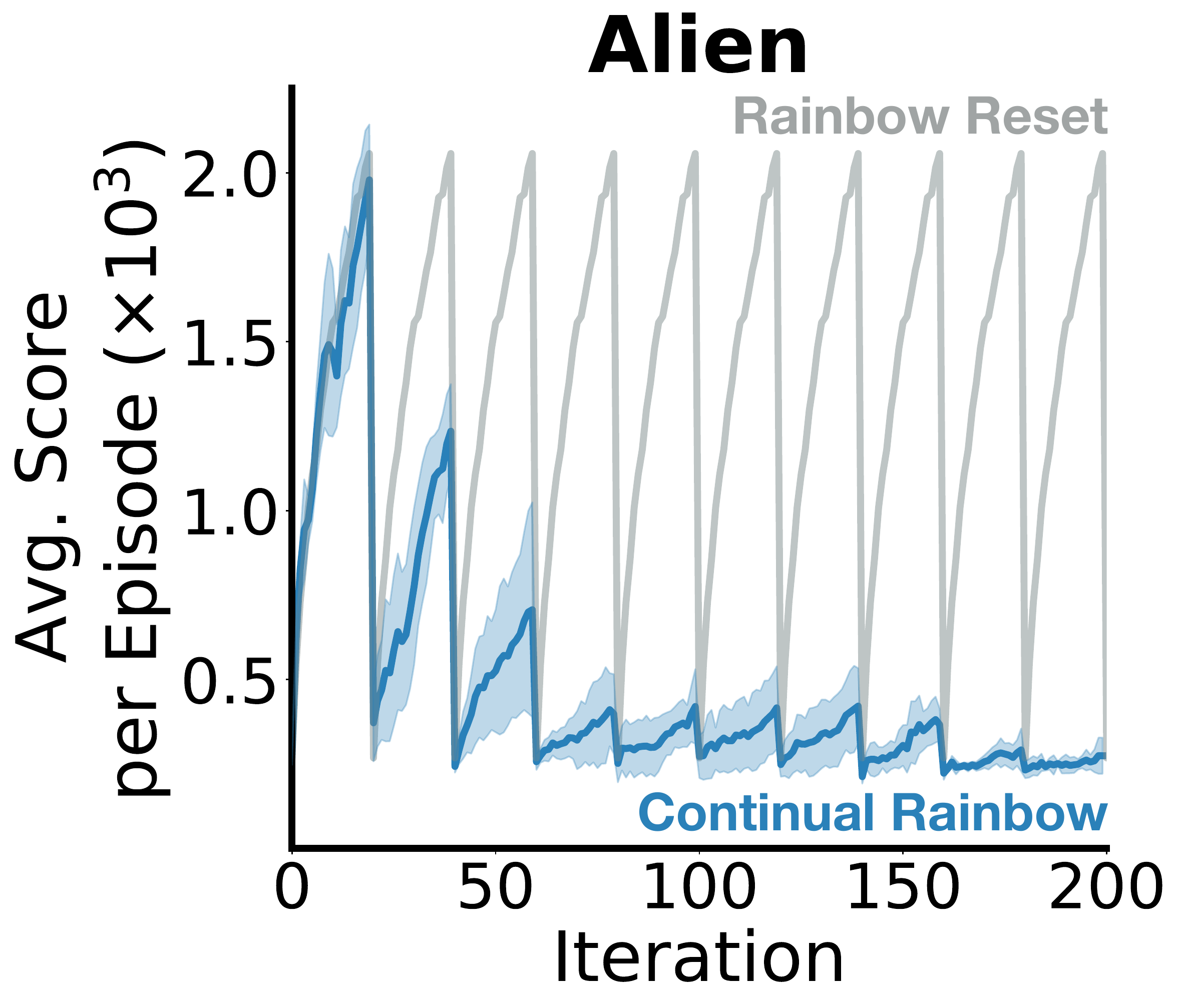}}
\caption{\textbf{Continual learning and the loss of plasticity in an Atari game:} online performance of a Rainbow agent (in blue) within a single game (Alien) when repeatedly playing a sequence of games.
Starting with Alien, the agent plays a sequence of 10 games, training on each game for 20M consecutive frames, and \revisits{} Alien after completing the cycle.
Rainbow's performance deteriorates on successive \visits{} of the game, indicating a loss in its ability to learn. 
For calibration, the gray line depicts the performance of an agent if it were reset after each switch, learning from scratch for 20M frames on each \visit{} to the game. The reset agent's performance would start low and reach the same level each time.
}
\label{fig:alien_loss_plasticity}
\end{center}
\end{wrapfigure}

Despite impressive benchmark performance of RL agents on simulated domains, the inherent non-stationarity of the reinforcement learning problem can still highlight limitations in the prevalent solution methods; limitations that are clearly visible on careful inspection.
For instance, neural networks lose capacity over time as they estimate continually changing value functions resulting from changing policies
~\citep{Kumar21Implicit, Lyle22Understanding}. Related, deep reinforcement learning systems \textit{overfit} to experiences gathered early in training, being unable to adjust as more experience becomes available---a phenomenon known as \textit{the primacy bias}~\citep{Nikishin22Primacy}. We hypothesize that these are all instances of the same underlying problem. Learning systems exhaust their learning capital over time and ultimately fail to meet the demands of a changing world, a phenomenon hereinafter referred to as {\em catastrophic loss of plasticity}~\citep{Dohare21Continual}.

In this paper, we contribute to the growing literature on catastrophic loss of plasticity in several ways.
First, we demonstrate catastrophic loss of plasticity in a performant deep reinforcement learning system, Rainbow~\citep{hessel2018rainbow}, on a variation of the Arcade Learning Environment (ALE)~\citep{bellemare2013arcade,machado2018revisiting} that involves non-stationarities from changing Atari 2600 games and game modes over time. Some of the experiments presented in this paper are run up to two billion environment interactions spanning 50 days of wall clock time.
The result depicted in Figure \ref{fig:alien_loss_plasticity} provides an example of the larger set of performance results presented in this paper.  
Next, we perform a careful analysis of weights, gradients, and activations sampled over the lifetime of the experiment. 
The analysis augments the existing findings in the literature and points to a remarkably simple mitigation. 
Finally, we evaluate one mitigation---the concatenated ReLU activation---and show its effectiveness.\\

\section{Preliminaries}

In this paper we study agents that continually interact with their environment and learn to achieve a goal. In particular, the agent's objective is to select actions based on some state, $S_t \in \mathcal{S}$, to maximize a scalar reward signal, $R_t \in \mathbb{R}$, over time. The agent's action, $A_t$, is determined by its policy, $\pi(A_t|S_t)$, which is adapted towards the goal of maximizing total discounted future reward: $G_t \doteq \sum_{i=0}^\infty\gamma^i R_{t+1+i}$, where $\gamma\in[0,1)$.

In this paper we focus exclusively on methods that learn value functions to obtain the policy: Deep Q-learning (DQN) and related variants. Put simply, a DQN agent continually adapts a parameteric estimate of future reward called the {\em value function}: $\hat{q}_{\bf w}(s, a) \approx \mathbb{E}[G_t|S_t=s, A_t=a]$, where ${\bf w}$ are the parameters of a neural network. On each timestep, DQN (1) selects an action according to some exploratory policy (e.g., $\epsilon$-greedy), (2) stores the most recent experience in replay buffer, and (3) samples a mini-batch from the replay buffer and updates the weights with gradients computed with the backpropagation algorithm: $\nabla_{\bf w}\mathbb{E}[(R_{t+1} - \gamma \max_{a\in\mathcal{A}}\hat{q}_{\bar {\bf w}}(S_{t+1},a )) - \hat{q}_{{\bf w}}(S_{t},A_t))^2]$, where $\bar {\bf w}$ are the parameters of the target network that are periodically set equal to ${\bf w}$. Often (e.g., in Atari 2600 games) the agent does not observe $S_t$, but instead an incomplete summary of the state such as the four most recent game frames.  \looseness=-1

The majority of our experiments focus on the Rainbow agent: an implementation based on DQN~\citep{Mnih2015Human}, which is highly tuned to play Atari 2600 games~\citep{hessel2018rainbow}. In particular, Rainbow combines several algorithmic components including distributional RL~\citep{bellemare2017distributional}, noisy networks for exploration~\citep{fortunato2017noisy}, prioritized experience replay~\citep{schaul2015prioritized}, dueling networks~\citep{wang2016dueling}, n-step returns~\citep{sutton2018reinforcement}, and a few other ideas. The performance of Rainbow is representative of the best performing model-free, synchronous (single environment) RL implementations available. Because of dueling networks, Rainbow's neural network has two streams, one that estimates the value function, $\hat{v}_{{\bf w_1}}(s) \approx \mathbb{E}[G_t | S_t = s]$, and one that estimates the advantage, $\hat{d}_{{\bf w_2}}(s,a)$, such that $\hat{v}_{{\bf w_1}}(s) + \hat{d}_{{\bf w_2}}(s,a) \approx \mathbb{E}[G_t | S_t = s, A_t = a]$. In this paper, we term the shared network layers, before the split for the two streams, \emph{convolution network}, the set of layers used in $\hat{v}_{{\bf w_1}}$'s stream \emph{value network}, and those used in $\hat{d}_{{\bf w_2}}$'s stream \emph{advantage network}.

In this paper we focus our empirical analysis on a subset  of Atari 2600 games using the Arcade Learning Environment (ALE)~\citep{bellemare2013arcade, machado2018revisiting}. The ALE is a standard domain for evaluating deep reinforcement learning algorithms and the two specific algorithms we study, Rainbow and DQN, were originally developed for Atari 2600 games.
Traditionally, the evaluation protocol within ALE consists of having an agent interact with a specific Atari 2600 game for 200 million frames, reporting the final performance achieved by the agent.
We abide by \citet{machado2018revisiting}’s recommendations of evaluation protocol, including the use of stochasticity, ignoring the lives signal, and reporting the average performance during training.
We diverge from the standard practice only by deploying a single agent, with a single neural network, to play a sequence of games and to do well in all of them (which we describe in detail in the next section).
In most prior work agents are forced to follow a specific schedule to ensure samples of all games are seen often during training \citep[e.g.,][]{reed2022generalist}. The setup explored in this paper is designed to mimic non-stationarity in continual learning tasks.

\section{Demonstrating Loss of Plasticity}
In this section we describe S-ALE, a continual variant of the ALE, and we use it to evaluate the performance of Rainbow under varying degrees of non-stationarity, demonstrating how the Rainbow agent loses plasticity over time.

\subsection{Adapting the ALE for Continual Learning}

Our objective is to evaluate how well deep RL methods can continually adapt in a changing world.
We consider a simple experimental regime that is derived from the popular ALE benchmark. 
The common practice in ALE is to develop performant algorithms that learn to play each game separately in a fixed number of interactions---typically 200M frames. 
Here, following the continual learning experimental practice of using a collection of tasks \citep{Zenke17Continual,thrun1995lifelong,taylor2009transfer,yu2020meta,khetarpal2022towards}, the agent learns on each game for a fixed number of frames and then switches to the next game \emph{without} resetting the learned weights nor the replay buffer contents in between.

Specifically, the environment cycles through a fixed sequence of games,  (e.g. Alien, Atlantis Boxing, ..., Alien, ...) for a fixed number of agent-environment interactions between any two switches (specifically 20M frames). Consequently, each game is repeated numerous times during the course of the experiment. 
We call the agent's experience playing a single game a \visit{}.
For instance, given a budget of 1 billion frames and a set of 10 games and 20M frames per game, it will take 200M steps to cycle through all games. The agent will get to play (\visit{}) each game 5 times over the length of the experiment, spending 180M frames between successive plays of the same game. Note, the duration of our experiments far exceeds the replay capacity of 1M frames, typical in ALE experiments. We refer to this experimental set up as Switching ALE, or S-ALE for short. 

\subsection{Learning Performance in S-ALE}

\begin{figure*}[tb]
\begin{center}
\centerline{\includegraphics[width=\textwidth]{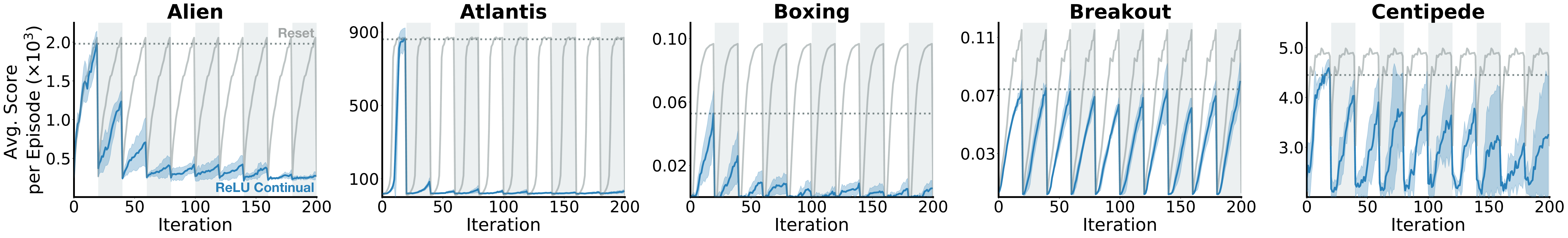}}
\caption{\textbf{Rainbow's loss of plasticity on a repeating sequence of 5 Atari 2600 games.} Each subplot reports learning within a single game. The blue curves show Rainbow learning as it cycles through 5 games in a fixed sequence. The agent interacts with each game for 20M frames in every iteration of the cycle; the cycle restarts every 100M frames. 
The gray line depicts the performance achieved by an idealized reset (described in text) that represents the performance of an agent that is reset every 20M frames; there would be no impact from other games or previous \visits{} to the current game. The dotted line marks the average performance achieved by the end of the first \visit{}---a visual reference to highlight performance drop over successive \visits{}.
\vspace{-.5cm}
}
\label{fig:loss_plasticity_5_games}
\end{center}
\end{figure*}

Figure~\ref{fig:loss_plasticity_5_games} depicts the performance of a Rainbow agent in S-ALE when cycling through 5 games, spending 20M frames per \visit{}. Prior work shows that Rainbow achieves impressive scores when trained on individual Atari 2600 games~\citep[c.f.][]{hessel2018rainbow}.
In the S-ALE setup, on the other hand, the agent performs relatively well when playing a game for the first time, but the performance dramatically decreases in most games on successive \visits{}.
We report results for each game in separate plots because the rewards in different games are not at the same scale. Moreover, the learning curve for each game (each subplot) reports only the performance of the agent while playing that game. The result for Alien, for example, only shows online learning performance when the agent is playing Alien; the agent's performance in other games is not in the same subplot.
Finally, each curve is averaged over five runs and the shading depicts their standard deviation.\footnote{Ideally, more runs would be used, but as mentioned above, these experiments took over a month to run. Consider these results demonstrative in nature: showing how deep RL agents can fail in a continual learning setup. Our results cannot provide concrete evidence that such failures will always occur.}
The results show that every time the agent \visit{}s a game again the agent learns more slowly than before, and the policy at the end of \visit{} is typically worse, ultimately collapsing to showing no signs of improvement.  

To help make it easier to contextualize these results consider an agent that completely forgets how to play a game after each switch, but maintains indefinitely its ability to learn the current game it is \visiting{}---relearning anew from scratch.
Furthermore, let us assume that the experience the agent has playing a game has no impact over its policy when learning another game (no interference, no generalization).
We call this the \emph{reset agent} as it equivalent to resetting the network weights, replay buffer, and optimizer state after every switch such that it is always learning from scratch.
This agent is far from ideal as it is forgetting absolutely everything it has experienced before, but because of its  never-ending learning potential, its performance will always recover in expectation to the same level seen at the end of the last \visit{}. Ideally any agent able to continually learn in the environment should be able to match and outperform the reset agent.
But the results in Figure~\ref{fig:loss_plasticity_5_games} show that Rainbow performs worse than the reset agent over time. 

We did not actually build and run a reset agent. Notice how all the bumps of the gray curve in Figure \ref{fig:loss_plasticity_5_games} are identical---one would expect some variation even if the agent was repeatedly reset. The results we report for the reset agent in this paper are idealized. For every game, we simply record the performance of a freshly initialized Rainbow on its first ever \visit{} to that game. Then we simply report the first-\visit{} performance for every successive \visit{} and label this as the {\em reset agent} in our plots.

We hypothesize that Rainbow is suffering from both catastrophic interference, the phenomenon where prior learning overrides previous learning \citep{french1999catastrophic,liu2020measuring}, and loss of plasticity, the phenomenon where the network loses the ability to learn. 
On \revisiting{} a game, we see that the agent not only needs to start relearning, it typically achieves much lower performance compared with the previous \visits{}.
Importantly, we observe this phenomenon consistently in different settings, as we vary the number of games the agent cycles through, or the frequency in which game switches happen. Figure~\ref{fig:loss_plasticity_10_games} (in the Appendix) shows a similar pattern when the agent cycles through 10 games. 
We also observe performance degradation when we change the number of frames per \visit{}. We see similar, if less pronounced results (as expected),  in the Appendix in Figure \ref{fig:loss_plasticity_10_games_10_iteration}, with 10M frames per \visit{}---thus many more \visits{} to each game over 1B frames. With 50M frames per \visit{} in Figure~\ref{fig:loss_plasticity_10_games_50_iteration}, also in the Appendix---only two \visits{} to each game over the experiment---we see similar degradation compared to the our first result with 20M frames per \visit{}. \looseness=-1    

Loss of plasticity is not unique to Rainbow, when cycling through 10 games with 20M and 50M frames per \visit{}, we also observe the same performance degradation with the DQN, as shown in Figure \ref{fig:dqn_loss_plasticity_10_games_20M} and \ref{fig:dqn_loss_plasticity_10_games_50M} in the Appendix. 

\subsection{Varying the Degree of Non-Stationarity: looping through game modes}

Here we consider a milder form of non-stationarity, induced by changing modes within a single game.
In the previous experiment we cycled through a sequence of games.
The degree of non-stationarity by changing games is presumably dramatic---the distribution of observations, the transition dynamics, even the scale of reward changes across games \citep[c.f.][]{machado2018revisiting,Farebrother18Generalization}. Indeed, there is some evidence of successful transfer of policies across Atari 2600 game modes \citep{Rusu22Probing,wang2022investigating}, hinting at a high degree of similarity. 
 
\begin{figure*}[tb]
\begin{center}
\centerline{\includegraphics[width=\textwidth]{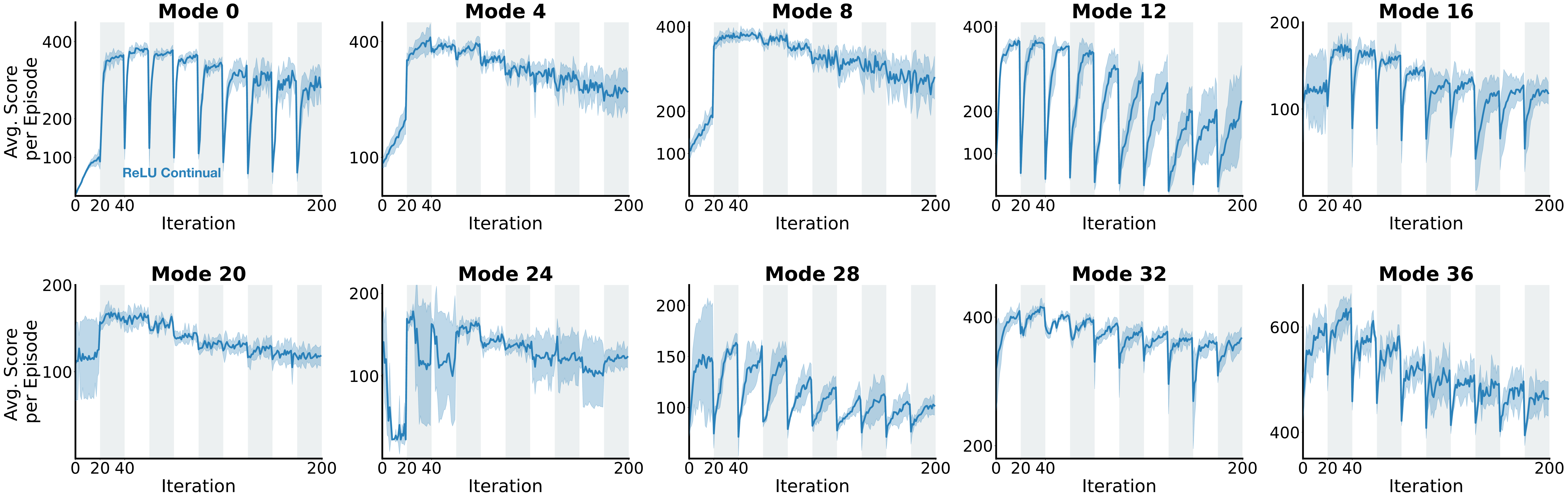}}
\caption{\textbf{Loss of plasticity in Breakout on a sequence of ten game modes.} The structure of this plot is similar to Figure \ref{fig:loss_plasticity_5_games}. The blue curves show Rainbow learning as it cycles through 10 different game modes of Breakout in a fixed sequence. The agent interacts with each game mode for 20M frames in every iteration of the cycle; the cycle restarts every 200M frames.
We again observe loss of plasticity in this sequence of modes despite the game modes being  similar to each other: similar dynamics and observation space (see the main text for details).}
\label{fig:breakout_10_modes}
\end{center}
\end{figure*}

We experimented with changing game modes in 3 Atari 2600 games (Breakout, Freeway, and Space Invaders) and, interestingly, we found clear evidence of loss of plasticity in only one.
In Breakout, Figure \ref{fig:breakout_10_modes}, the performance deteriorates over successive \visits{} to the same game mode, paralleling the result of changing games. Looking closely at Figure \ref{fig:breakout_10_modes}, the most dramatic performance loss is observed in game modes 4, 8, 20, and 36. In Breakout, the agent must learn to reflect the ball off a paddle which it can only move left or right along the bottom of the screen. In game mode 8, the agent has two additional actions: catch and release the ball, which affords significantly more control. Similarly, modes 4, 30, and 36 allow the agent to steer the ball after reflection off the paddle. These four modes present the biggest changes in dynamics, potentially explaining why the performance in these modes matches our first result (cycling through different games).

In Freeway (Figure \ref{fig:freeway_8_modes}), the agent recovers faster when \revisiting{} a game mode and exceeds performance of prior \visits{}.
Finally, in Space Invaders (Figure \ref{fig:space_invaders_10_modes}), the agent continues to perform well across \visits{} to the same game mode with the exception of one game mode.  All figures referenced in this paragraph are in the Appendix. These results suggest that loss of plasticity is somewhat related to the similarity between the different tasks an agent might be faced with, but it is difficult to characterize such similarities beforehand. For all Atari 2600 games investigated the different game modes induce, one way or another, a change in dynamics, but specific ones (as in Breakout) seem to have a bigger impact than others.

\section{Characterizing Loss of Plasticity}\label{sec:loss_of_plasticity}

\begin{figure*}[tb]
\vskip 0.2in
\begin{center}
\centerline{\includegraphics[width=\textwidth]{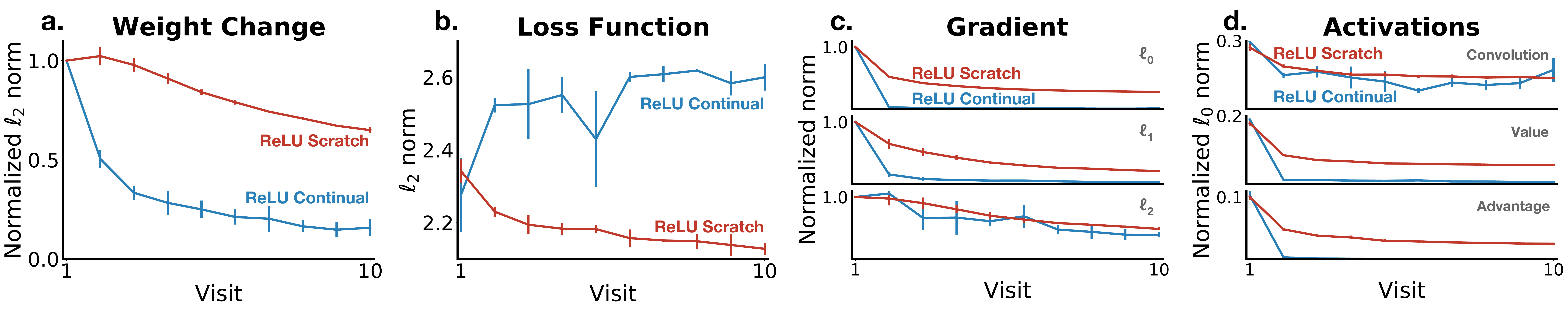}}
\caption{\textbf{Weight Change in Alien.} Normalized weight change of the Rainbow network over successive plays of Alien. \textbf{Loss Function.} Loss at halfway point over successive plays. For calibration, we include a scratch baseline plays Alien exclusively and uninterrupted and sees the same number of game frames (see text for details). \textbf{Gradient.} Normalized gradient norms aggregated across layers ($\ell_2$, $\ell_1$, and $\ell_0$). \textbf{Activations}. $\ell_0$ norm of the activations.}
\label{fig:characteristics}
\end{center}
\vskip -0.2in
\end{figure*}

In this section we take a closer look at how the internal components of Rainbow evolve over time.
Before we start, it is helpful to summarize our findings. 
As shown earlier, the performance progressively worsens each time the Rainbow agent \revisits{} a game and, in some cases, it collapses altogether.
We observe the first hint of reduced plasticity by inspecting how much the network weights change each time the agent \visits{} a game---in the game Alien, for example, the {\em weight change} consistently shrinks with additional \visits{} (c.f. Figure \ref{fig:characteristics}{\color{red} a}). 
More critically, the weight change continues to diminish even though the (distributional) loss remains large (Figure~\ref{fig:characteristics}{\color{red} b}). 
One would expect a large loss to result in large gradients, but we find that the magnitude of gradient also diminishes on every successive \visit{} (Figure~\ref{fig:characteristics}{\color{red} c}).
Finally, looking at the activations over time, we see fewer and fewer units are active (Figure~\ref{fig:characteristics}{\color{red} d}). This makes sense. The Rainbow agent uses Rectified Linear Units (ReLUs). If a ReLU is not active (i.e. zero output value), then it does not contribute to the change in its incoming weights (via the chain rule).
A learning system that is made up of unchanging weights, irrespective of a large global loss, can be well thought of as having lost its ability to learn new things---the network is losing plasticity.

Next we discuss how we computed each of the statistics discussed above and we present more detailed comparisons.

\subsection{Weight Change Diminishes Over Time}
We compare the weight change over every \visit{} in S-ALE to the weight change produced by an agent learning from scratch in a conventional (non-continual) ALE experiment. 
Specifically, we compute the $\ell_2$-norm of the weight change for every \visit{} from when the agent start playing Alien until it is halfway through the current \visit{} (i.e. over 10M frames). We compute the weight change for the {\em scratch agent} by simply stacking \visits{} one after the other and no other game is played in between. Results are shown in Figure \ref{fig:characteristics}{\color{red}a}. 

To make this comparison meaningful, because Rainbow uses a multi-layered network and the weight magnitudes across layers are not scale invariant, we normalize the weight changes. 
We use the weight change observed in the first \visit{} to scale the norms of weight change in subsequent \visits{} (thus the scaled norm can be greater than one). 
We aggregate the result from all layers with a weighted average, weighted by the number of weights in that layer.

We contrast weight change with how the loss changes during training. We compute the loss depicted in Figure \ref{fig:characteristics}{\color{red}b} by averaging the loss halfway through each \visit{}: halfway through each \visit{}, we average the empirical loss on 100 mini-batches. 
In the continual set up, the weights change diminishes sharply---on the tenth \visit{}, the weight change is 20\% of that in the first \visit{} (blue curve in Figure \ref{fig:characteristics}{\color{red}a}).
This is notable because the loss, on the other hand, grows large.  
In contrast, the scratch agent's weights (the red curve in Figure \ref{fig:characteristics}{\color{red}a}) continue to change much more significantly: the weight change on the tenth \visit{} is still 75\% of that in the first \visit{} (consistent with the decreasing loss for the scratch agent). \looseness=-1

\subsection{Gradient Collapse}
The norm of the gradients helps us understand why the weights stop changing, but the loss remains high.
We summarize the gradients as follows: at the halfway point of each \visit{}, we compute the average layer-wise norm of the gradients for the next 100 updates. 
We employ the same normalization and aggregation approach as the weight change result.
Figure \ref{fig:characteristics}{\color{red}c} reports the $\ell_0$, $\ell_1$, and $\ell_2$ norms of the gradients for the continual and scratch agents. 
The $\ell_0$ and $\ell_1$ norms of the gradients for the continual set up (blue curves) decay much faster than the corresponding norms of the scratch agent (red curves). The gradients inside Rainbow's network collapse almost to zero in S-ALE! Note the decay in $\ell_2$ is not as drastic, likely because $\ell_2$ norm's susceptibility to outliers due to squaring.

\subsection{Activation Collapse}

Rainbow exhibits poor performance, increasing training loss, reducing weight change, and dying gradients---inspecting the activations in the network shed some light on the collapse of gradients and it also points toward a solution. 
At halfway point of each \visit{}, we record the internal activations of the network for the same 100  mini-batches that were used for the gradient inspection.  
The network in the Rainbow agent consists of a stack of convolution layers whose output is fed into a value network and an advantage network.
The output of the convolution stack consists of 3136 units; the value and the advantage network have one hidden layer each of size 512.
In Figure \ref{fig:characteristics}{\color{red} d}, we show the $\ell_0$ norm of each layer's activations scaled by the number of units, average across 100 mini-batches of size 32.
We see that the $\ell_0$ norm of activations in the advantage and value network collapses: a very small fraction of the units of the Rainbow's network in the continual setting produce non-zero value.
If a ReLU is zero, then it does not contribute to the change in its incoming weights, hindering adaptation. 
Though activations of the convolution stack do not collapse, the adaptation of convolution layers relies on the gradients to flow back from the advantage and value layers---consequently, task changes in S-ALE will impact the convolutional layers less and less over time.

\section{Mitigating loss of plasticity with Concatenated ReLUs}
In this section we explore a simple approach to increase and maintain network activations and thus mitigate the loss of plasticity. 
Recall that the internal activations of Rainbow's network decayed to the point where only a small fraction of the units produced non-zero values (less than 1\% in fact).
This collapse in activation resulted in diminishing gradients and prevented further change to the weights, thus, impeding learning.
Here we consider a simple change to the network architecture that can prevent activation collapse: Concatenated Rectified Linear Units (CReLUs).
Originally proposed to understand and improve convolutional architectures for object recognition \citep{Shang16Understanding}, CReLU concatenates the input with its negation and applies ReLU to the result: $\text{CReLU}(x) \doteq [ReLU(x), ReLU(-x)]$. CReLU outputs twice the number of input signals and ensures that, for every input signal, one of the two output signal is nonzero, except when the input signal is exactly zero.

\begin{figure*}[tb]
\begin{center}
\centerline{\includegraphics[width=\textwidth]{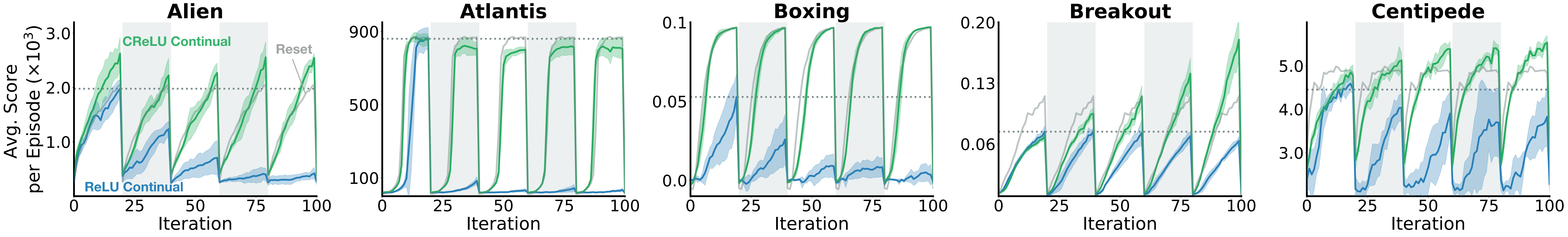}}
\caption{\textbf{CReLU activation function mitigates Rainbow's loss of plasticity in S-ALE.} The structure of this plot is identical to Figure~\ref{fig:loss_plasticity_5_games} The blue curves show the performance of the Rainbow agent with ReLU activations in the continual learning setting as it cycles through 5 different Atari 2600 games in a fixed sequence. The agent interacts with each game for 20M frames in every iteration of the cycle; the cycle restarts every 100M frames. The green curve shows the performance of Rainbow with CReLU activations in the S-ALE experimental setup. 
CReLUs typically do not result in faster initial learning (when compared to the reset agent) but maintains the ability to learn, continuing to match (or exceed, as in Boxing) the performance of reset agent and its own performance on prior \visits{}---a basic requirement for continual learning.}
\label{fig:crelus_dh_5_games}
\end{center}
\end{figure*}

Evaluating CReLUs requires us to pay attention to the effective capacity of the network. 
In a given layer, replacing ReLU with CReLU doubles the number of activations while keeping the number of parameters the same. This doubling can potentially double the number of parameters in the following layer.
To isolate the contribution of CReLU in mitigating the loss of plasticity, we experiment with two approaches.
The first approach controls for the number of input signals before the application of the activation function (\textit{invariant input dimension}).
For the CReLU variant of continual Rainbow, this doubles the number of parameters in every layer (except the first one).
The second approach controls for the number of output signals after the application of the activation function (\textit{invariant output dimension}). 
This halves the number of parameters in every layer (except the last one) in the CReLU variant.
We evaluate both architectures in our S-ALE setup with no change in hyper-parameters. 

The performance of the Rainbow agent with CReLU and \textit{invariant input dimension} is depicted in Figure~\ref{fig:crelus_dh_5_games}.
The results show that Rainbow-CReLU---unlike Rainbow-ReLU---continues to maintain or improve its performance on successive \visits{} in all five games.
On every repeated \visit{} to a game, having forgotten what was learned on the previous \visit{}, Rainbow-CReLU (like Rainbow with ReLUs) must  learn to play the game again.
The Rainbow-CReLU agent demonstrates improved plasticity by matching (or exceeding, in Breakout and Centipede, for example) the performance of the reset agent and its own performance in prior \visits{}. The results for the invariant output dimension can be seen in Figure \ref{fig:crelus_size} in the Appendix. This variant of Rainbow-CReLU with less parameters also clearly demonstrates maintenance of plasticity.

\begin{figure*}[tb]
\begin{center}
\centerline{\includegraphics[width=\textwidth]{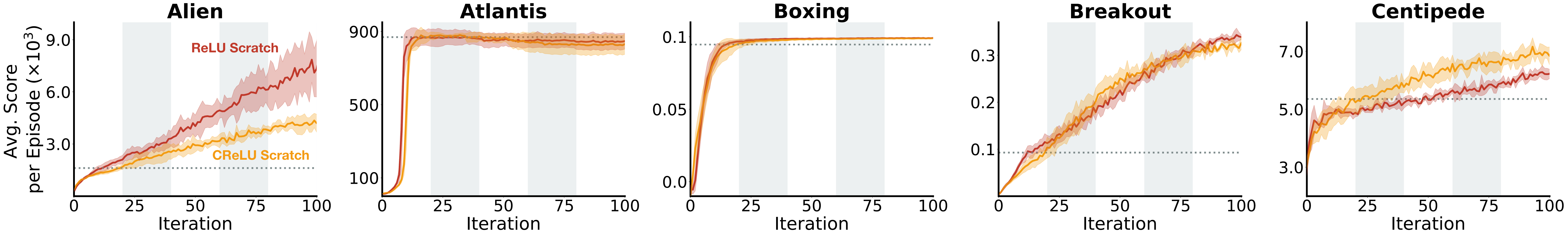}}
\caption{\textbf{Comparing Rainbow with CReLUs and Rainbow with ReLUs.} 
When training a Rainbow agent from scratch in a single game for 100M frames, CReLU and ReLU activations lead to comparable performance.}
\label{fig:comparison_relu_crelu}
\end{center}
\end{figure*}

It would be reasonable to suspect that the improvement in plasticity is simply due to the general superiority of CReLU over ReLU, but this does not seem to be the case. Figure \ref{fig:comparison_relu_crelu} shows that ReLU-Rainbow (i.e. ReLU Scratch) performs comparably in the conventional (non-continual) ALE, that is, when only learning from scratch in each game. For completeness, we also compared continual Rainbow-CReLU against the scratch Rainbow-CReLU agent (as opposed to the scratch Rainbow-ReLU in the plots above) to quantify the impact in performance that cycling through multiple games instead of learning only from scratch has when using CReLU activations. The result can be found in Figure \ref{fig:continual_learning_only_with_crelus_5_games} in the Appendix.

\begin{figure*}[tb]
\begin{center}
\centerline{\includegraphics[width=\textwidth]{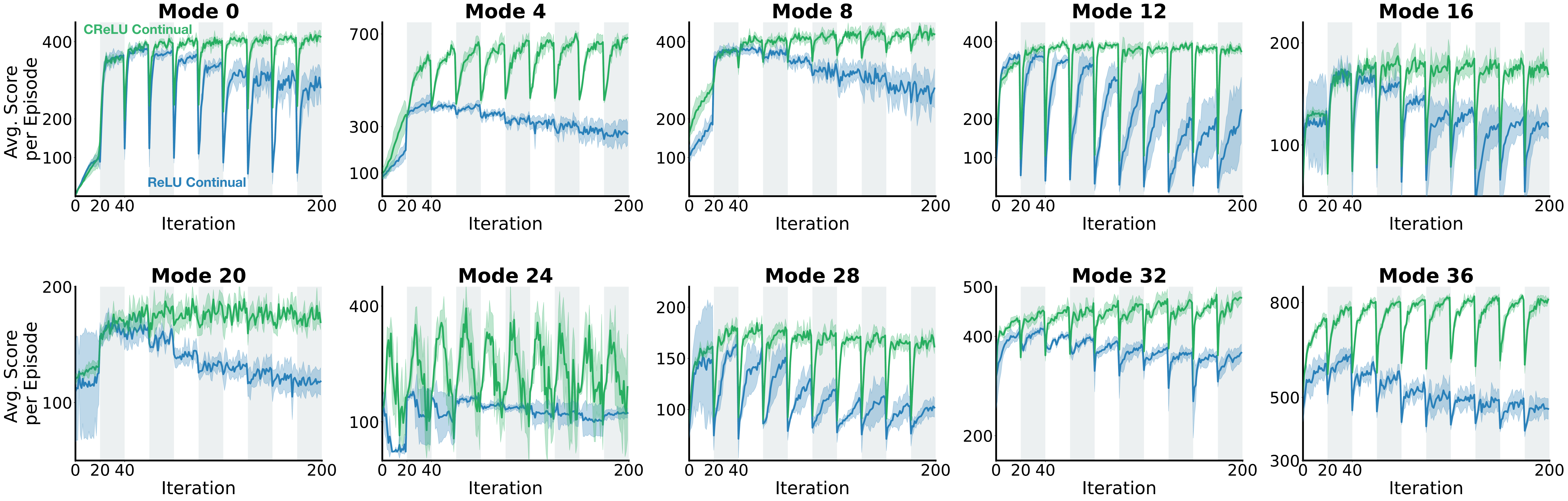}}
\caption{\textbf{Mitigating loss of plasticity in Breakout game modes with CReLU activations.} The structure of this plot is identical to Figure~\ref{fig:breakout_10_modes}. The blue curves show the performance of the Rainbow agent with ReLU activations in the continual learning setting as it cycles through 10 different game modes of Breakout in a fixed sequence. The agent interacts with each game mode for 20M frames in every iteration of the cycle; the cycle restarts every 200M frames. The green curves show the performance of Rainbow with CReLU activations in the same continual learning setting.}
\label{fig:breakout_10_modes_crelu}
\end{center}
\end{figure*}

We see a similar pattern when we evaluate Rainbow-CReLU for sequential training on Breakout game modes. As depicted in Figure~\ref{fig:breakout_10_modes_crelu}, Rainbow-CReLU continues to maintain or improve its performance on successive \visits{} in all 10 game modes. Rainbow-CReLU shows no significant improvement over Rainbow across the game modes of Space Invaders, presumably because Rainbow did not experience significant loss of plasticity (see Figure \ref{fig:space_invaders_10_modes_crelu} in the Appendix for the evaluation of Rainbow-CReLU in the different game modes of Space Invaders).

\begin{figure*}[tb]
\vskip 0.2in
\begin{center}
\centerline{\includegraphics[width=\textwidth]{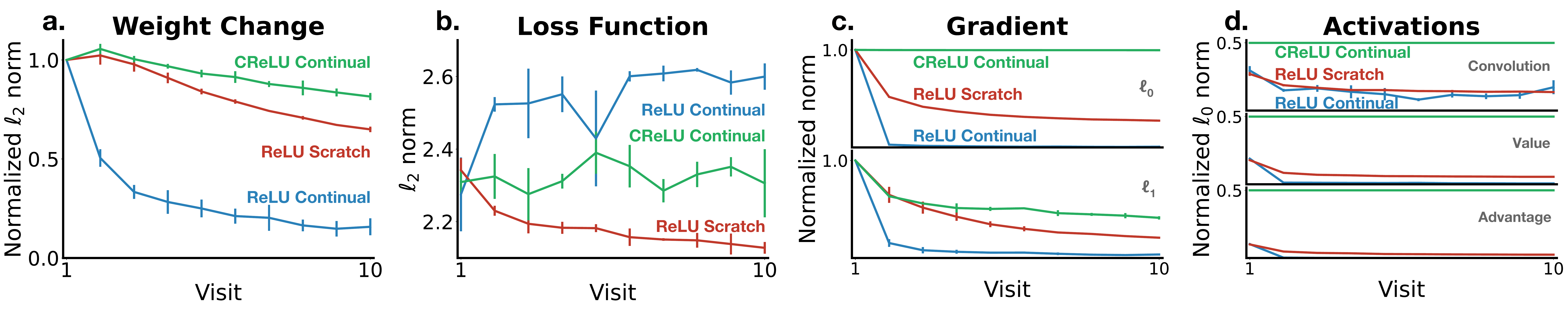}}
\caption{\textbf{Weight Change in Alien.} Normalized weight change of the Rainbow network over successive plays of Alien. \textbf{Loss Function.} Loss at halfway point over successive plays. For calibration, we include a scratch baseline that plays Alien exclusively and uninterrupted and sees the same number of game frames (see text for details). \textbf{Gradient.} Normalized gradient norms aggregated across layers ($\ell_2$, $\ell_1$, and $\ell_0$). \textbf{Activations}. $\ell_0$ norm of the activations.}
\label{fig:characteristics_crelu}
\end{center}
\vskip -0.2in
\end{figure*}

Finally, we inspect the internal statistics of the Rainbow-CReLU agent to validate whether they are consistent to what we would expect based on the discussion in Section~\ref{sec:loss_of_plasticity}. They constitute clear evidence of improvement in plasticity. As before we focus on the game Alien.
First, Figure \ref{fig:characteristics_crelu}{\color{red}d} shows that about half of the activations continue to produce non-zero values; CReLU is preventing activation collapse (the green curves).
It does so by construction: for every input signal, one of the two output signals is nonzero, except when the input signal is exactly zero (unlikely because of the 32-bit floating point representation of real numbers).
Figure \ref{fig:characteristics_crelu}{\color{red}c} shows that the relative norm of the gradients stays large (as we would expect if activations did not collapse). The $\ell_0$ norm does not diminish relative to the first \visit{}. Relative $\ell_1$ norm does reduce, however, it does not collapse like with ReLUs, reducing much more slowly than Scratch; and the rate of decay becomes smaller over \visits{}. 
Figure \ref{fig:characteristics_crelu}{\color{red}a} shows that weights in Rainbow-CReLU continue to change significantly relative to the first \visit{} all the way through ten \visits{}: as we would expect if gradients did not collapse. The relative weight change for Rainbow-CReLU is larger than not only Rainbow-ReLU but also the Scratch baseline, which makes sense given that the loss function observed with CReLUs is larger than the loss function observed with ReLU Scratch. Figure~\ref{fig:characteristics_crelu}{\color{red}b} shows the sampled loss of Rainbow-CReLU. It does not grow to be as large as Rainbow-ReLU, likely due to its improved plasticity that allows it to continue to learn. Note, as aforementioned, that the training loss of continual Rainbow with CReLUs is still higher than Scratch; this is the topic of the next section.

\clearpage

\section{Catastrophic Forgetting: an unresolved challenge in continual learning}
Up to this point we have discussed a large number of results, and thus it is useful to summarize what we have learned.
We have focused on catastrophic loss of plasticity as the core phenomenon, demonstrating it in the S-ALE set up. We analyzed the learning dynamics over time and evaluated a simple mitigation strategy.
In this section, we discuss how the same S-ALE experiments relate to the well-established phenomenon of catastrophic forgetting ~\citep{french1999catastrophic}. \looseness=-1 

\begin{figure*}[!h]
\begin{center}
\centerline{\includegraphics[width=\textwidth]{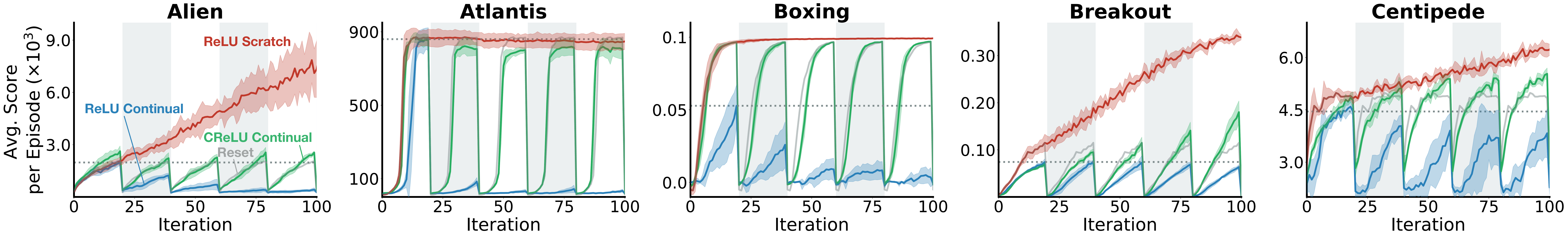}}
\caption{\textbf{Catastrophic forgetting and the loss of plasticity in S-ALE.} The structure of this plot is identical to Figure~\ref{fig:crelus_dh_5_games}. Here we also include in red the performance of a conventional Rainbow agent (with ReLU activations) training on each game, individually and uninterrupted, for 200M frames---the conventional non-continual ALE setup. We expect no notable loss of plasticity in this setting, and thus the red curve sets a clear performance bar for the other agents. This provides clear context for the CReLU result:
for performance within a single game, there is no benefit from learning the other games, and barring the two games in which good performance can be reached within a single visit (Atlantis and Boxing), the gulf between the performance of scratch baseline and continual CReLU gets wider over successive visits---a sign of catastrophic interference. 
}
\label{fig:complete_plot_5_games}
\end{center}
\end{figure*}

In all the results discussed in this paper, every time the Rainbow agent \revisits{} a game (or mode), it shows no retention of what was previously learned: the agent forgets.
The CReLU mitigation does not address this issue and we see no reason to expect as such.
Consider the results in Figure \ref{fig:complete_plot_5_games}. 
The Rainbow agent with CReLU activations, when cycling through five games, can improve the performance over successive \visits{} in some cases (green curve), but the improvement appears modest when we compare with an agent that learns uninterrupted on a single game with the same amount of experience (red curve). Figure \ref{loss_plasticity_10_games_final} in the Appendix shows similar results for a ten-game sequence and Figure~\ref{fig:continual_learning_only_with_crelus_5_games}, also in the Appendix, presents similar results when using CReLU Scratch as a baseline.

It is unclear if it is realistic to expect agents such as Rainbow---designed for non-continual ALE---to remember what they learned over 100M frames ago and transfer learning between \visits{}. Nonetheless, there is a widening gulf between the idealized agent that learn uninterrupted (the red curve in Figure \ref{fig:complete_plot_5_games}) and continual Rainbow. The path forward seems clear: synthesizing continual learning approaches that address stability-plasticity dilemma and address both catastrophic forgetting and the loss of plasticity. 

\clearpage

\section{Conclusions and Future Work}

Continual adaptation is essential as no amount of prior learning is sufficient in complex and continually changing environments.
Most of the solution methods in deep RL have been developed and evaluated in mostly stationary settings, and it is often assumed that these solution methods are naturally applicable to continual learning. 
In this paper we showed that canonical value-based deep RL methods in DQN and Rainbow are in fact not able to perform well in continual learning problems.

The experiments were performed in a non-stationary setting where the agent is tasked to play a sequence of Atari 2600 games without any resetting in between.
The results demonstrated that deep RL algorithms can lose their ability to learn (i.e. update the weights of the neural network), a phenomenon known as the loss of plasticity.
We inspected the evolution of weights, gradients, and activations over the course of learning and observed activation collapse: in the continual setup with a non-stationary environment, a tiny fraction of network units produce non-zero activation value, inhibiting adaptation. 
Finally, we demonstrated that swapping ReLUs with CReLUs mitigate the loss of plasticity. 

There are several promising directions for future work.
Further unifying the observed phenomenon in S-ALE to related observations--e.g., implicit under-parameterization, primacy bias, and capacity loss---can lead to better tools for instrumenting deep RL methods.
Of course, CReLUs are not the only way of tackling the issues identified in this paper.
For example, re-initializing parts of the network randomly, or according to some notion of utility~\citep[e.g.,][]{Dohare21Continual} are promising solutions that should be further investigated.
More importantly, while CReLUs do mitigate the loss of plasticity, they are not sufficient for the agent to effectively reuse past experience---they do not resolve catastrophic interference. This paper represents a small step toward the goal continual deep reinforcement learning: developing agents that can maintain plasticity in the face of an ever changing world and efficiently leverage prior learning.

\section*{Acknowledgements}
This work was developed while the authors were at DeepMind. The authors would like to thank Thomas Degris for his thorough feedback on an earlier draft, and Joshua Davidson, Finbarr Timbers, and Doina Precup for useful discussions.

\bibliography{example_paper}
\bibliographystyle{collas2023_conference}

\newpage
\appendix
\section{Appendix}

\begin{figure*}[htb!]
\begin{center}
\centerline{\includegraphics[width=\textwidth]{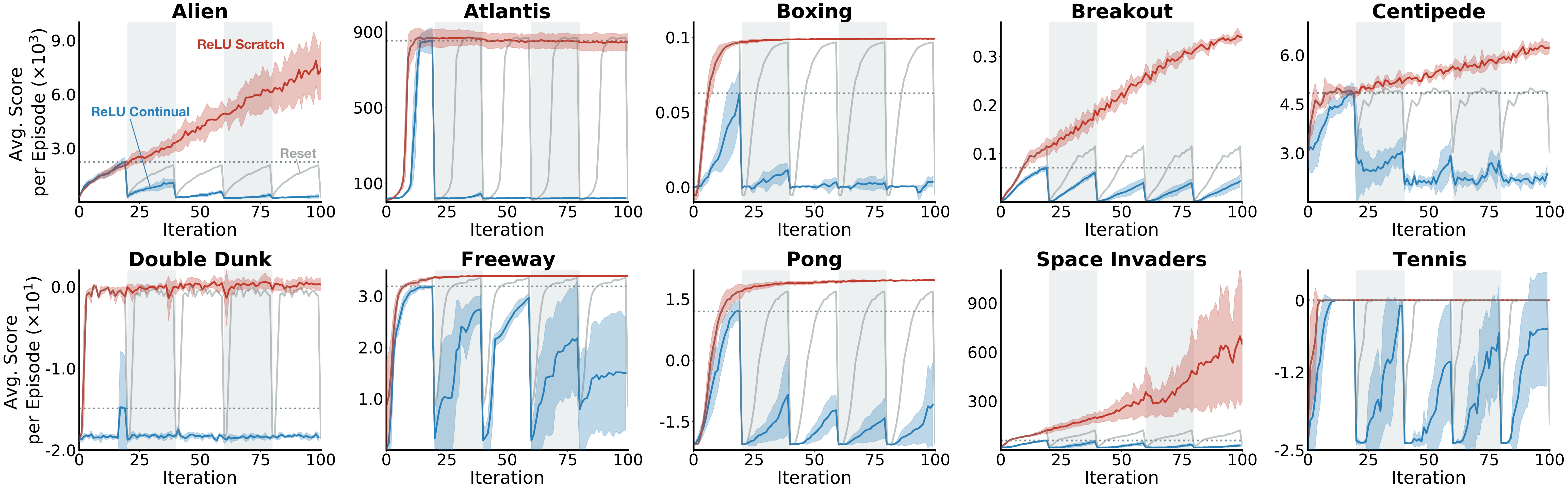}}
\caption{\textbf{Rainbow's Loss of Plasticity on a repeating sequence of 10 Atari 2600 games with 20M steps on each game.} Each subplot reports learning within a single game. The blue curves show the Rainbow learning as it cycles through 10 games in a fixed sequence. The agent interacts with each game for 20M frames in every iteration of the cycle; the cycle restarts every 200M frames. The gray line depicts the performance achieved by an idealized reset (described in text) that represents the performance of an agent that is reset every 20M frames; there would be no impact from other games or previous visits to the current game. The dotted line marks the average performance achieved by the end of the first \visit{}---a visual reference to highlight performance drop over successive \visits{}. Rainbow's performance when learning from scratch in a single game for 200M frames is represented in red. The red curve can be seen as an approximation to the idealized performance of an agent that does not suffer from interference (nor benefits from generalization) from other modes, and that does not forget anything about its past experiences, restarting where it left off in the previous \visit{}.}
\label{fig:loss_plasticity_10_games}
\end{center}
\end{figure*}

\begin{figure*}[htb!]
\begin{center}
\centerline{\includegraphics[width=\textwidth]{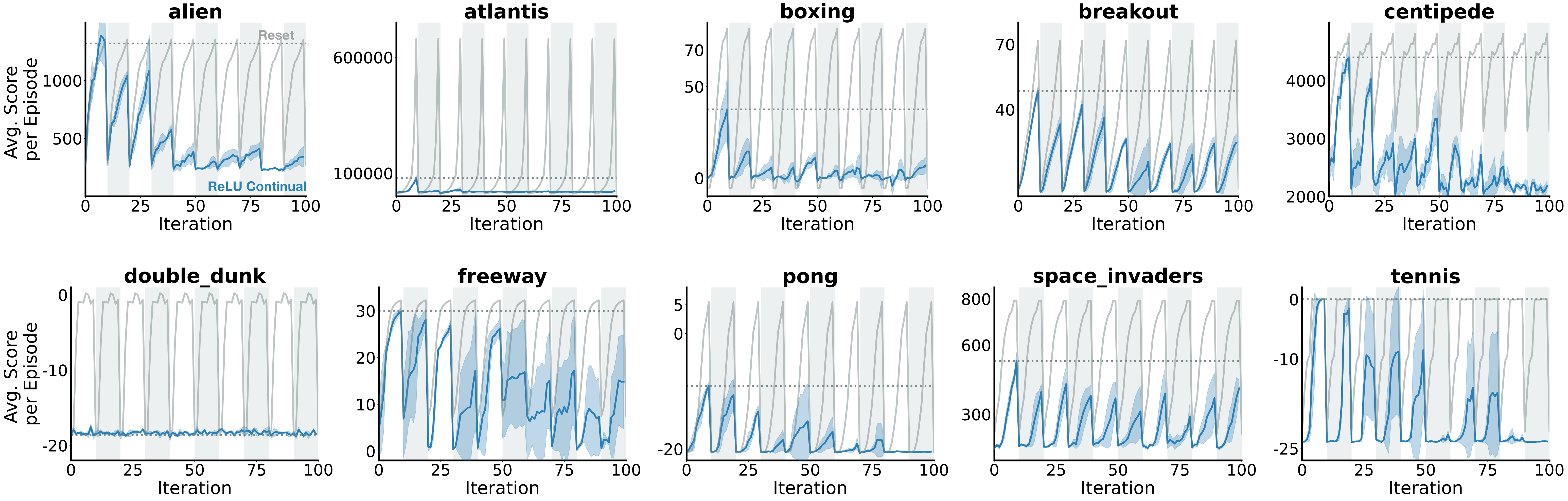}}
\caption{\textbf{Rainbow's performance on a repeating sequence of 10 Atari games with 10M frames between game switches.} Each subplot reports learning within a single game. The blue curves show Rainbow learning as it cycles through 10 games in a fixed sequence. The agent interacts with each game for 10M frames in every iteration of the cycle; the cycle restarts every 100M frames. The gray line depicts the performance achieved by an idealized reset (described in text) that represents the performance of an agent that is reset every 10M frames; there would be no impact from other games or previous visits to the current game. The dotted line marks the average performance achieved by the end of the first \visit{}---a visual reference to highlight performance drop over successive \visits{}.}
\label{fig:loss_plasticity_10_games_10_iteration}
\end{center}
\end{figure*}

\begin{figure*}[htb!]
\begin{center}
\centerline{\includegraphics[width=\textwidth]{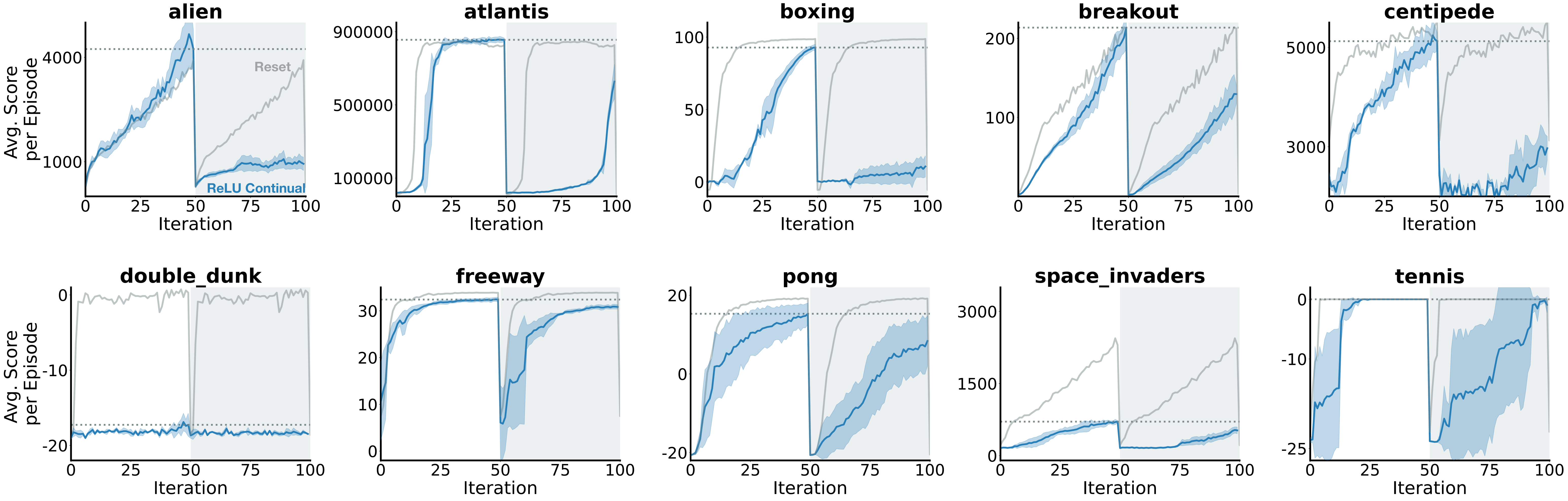}}
\caption{\textbf{Rainbow's performance on a repeating sequence of 10 Atari games with 50M frames between game switches.} Each subplot reports learning within a single game. The blue curves show Rainbow learning as it cycles through 10 games in a fixed sequence. The agent interacts with each game for 50M frames in every iteration of the cycle; the cycle restarts every 500M frames. The gray line depicts the performance achieved by an idealized reset (described in text) that represents the performance of an agent that is reset every 50M frames; there would be no impact from other games or previous visits to the current game. The dotted line marks the average performance achieved by the end of the first \visit{}---a visual reference to highlight performance drop over successive \visits{}.}
\label{fig:loss_plasticity_10_games_50_iteration}
\end{center}
\end{figure*}

\begin{figure*}[!htb]
\begin{center}
\centerline{\includegraphics[width=\textwidth]{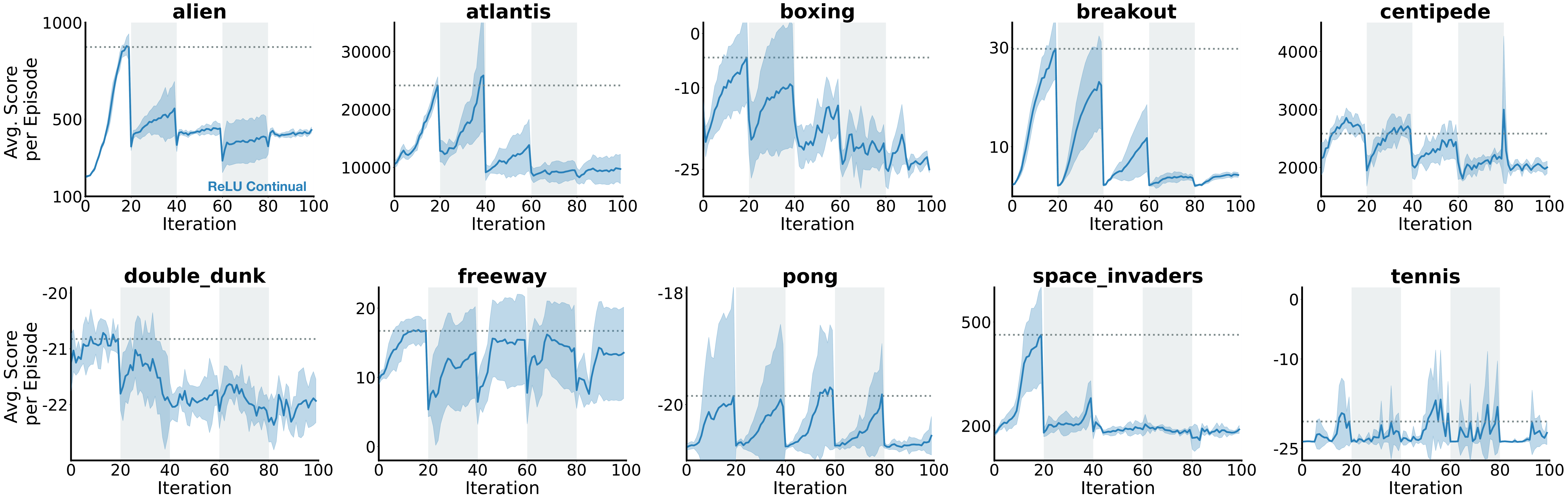}}
\caption{\textbf{DQN's Loss of Plasticity on a repeating sequence of 10 Atari 2600 games with 20M steps on each game.} Each subplot reports learning within a single game. The blue curves show DQN learning as it cycles through 10 games in a fixed sequence. The agent interacts with each game for 20M frames in every iteration of the cycle; the cycle restarts every 200M frames. 
The dotted line marks the average performance achieved by the end of the first \visit{}---a visual reference to highlight performance drop over successive \visits{}.}
\label{fig:dqn_loss_plasticity_10_games_20M}
\end{center}
\end{figure*}

\begin{figure*}[!htb]
\begin{center}
\centerline{\includegraphics[width=\textwidth]{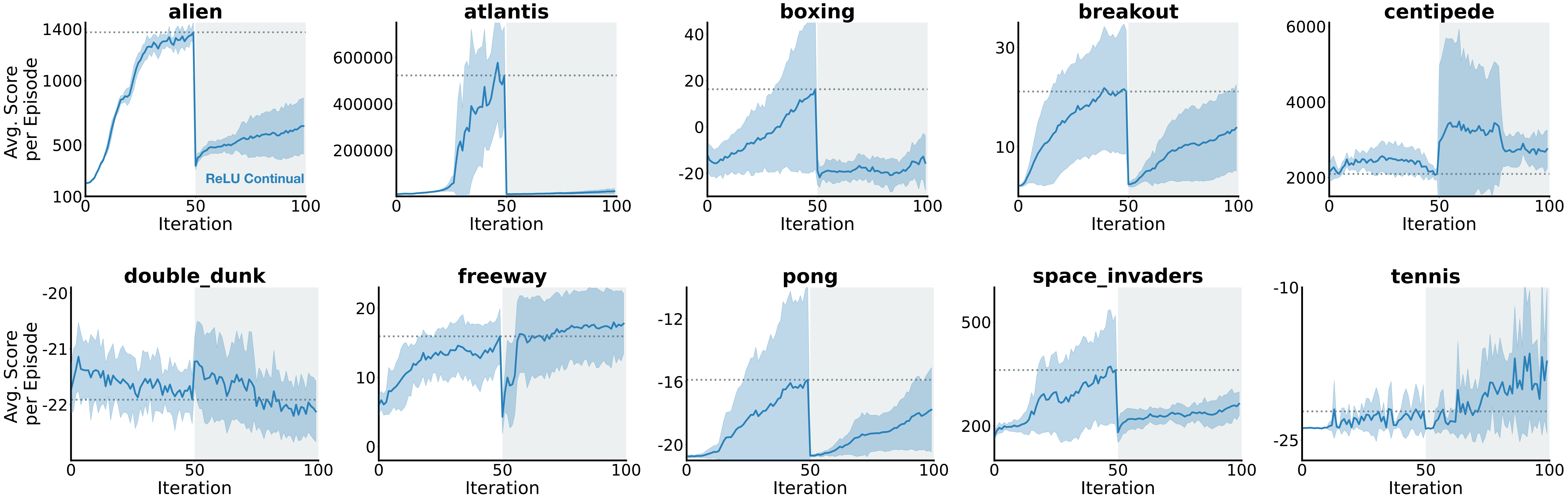}}
\caption{\textbf{DQN's Loss of Plasticity on a repeating sequence of 10 Atari 2600 games with 50M steps on each game.} Each subplot reports learning within a single game. The blue curves show DQN learning as it cycles through 10 games in a fixed sequence. The agent interacts with each game for 50M frames in every iteration of the cycle; the cycle restarts every 500M frames. The dotted line marks the average performance achieved by the end of the first \visit{}---a visual reference to highlight performance drop over successive \visits{}.}
\label{fig:dqn_loss_plasticity_10_games_50M}
\end{center}
\end{figure*}

\begin{figure*}[htb!]
\begin{center}
\centerline{\includegraphics[width=\textwidth]{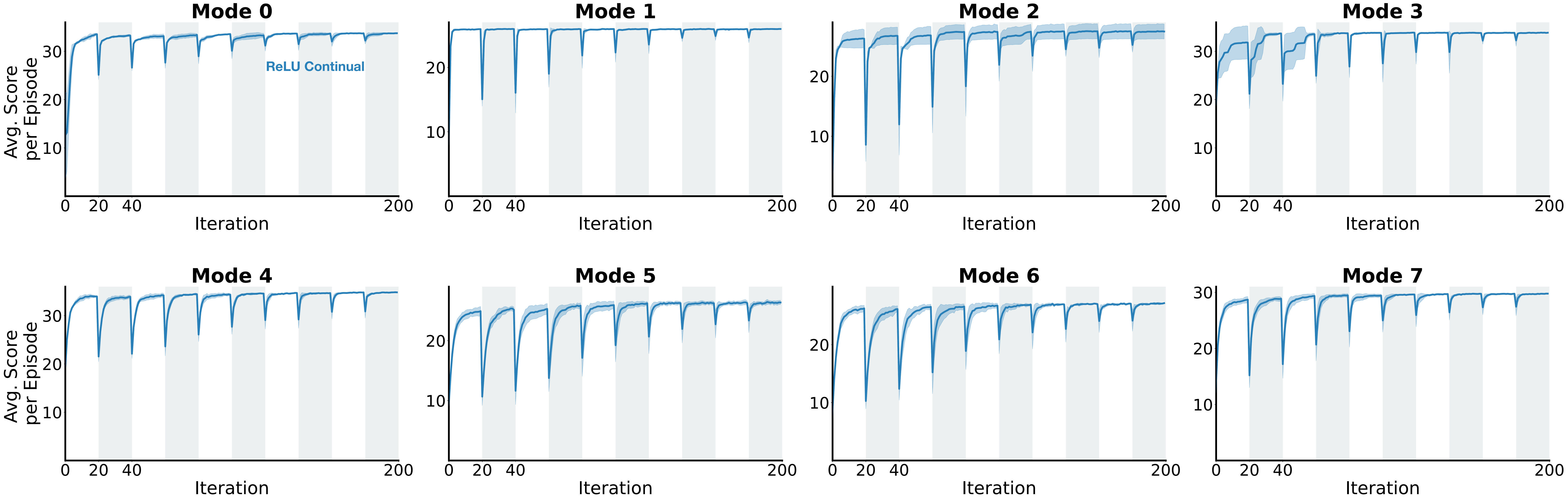}}
\caption{\textbf{Loss of Plasticity was not observed in Freeway on a sequence of 10 game modes.} The structure of this plot is identical to Figure~\ref{fig:breakout_10_modes}. The blue curves show Rainbow learning as it cycles through 8 different game modes of Freeway in a fixed sequence. The agent interacts with each game mode for 20M frames in every iteration of the cycle; the cycle restarts every 160M frames.}
\label{fig:freeway_8_modes}
\end{center}
\end{figure*}

\begin{figure*}[htb!]
\begin{center}
\centerline{\includegraphics[width=\textwidth]{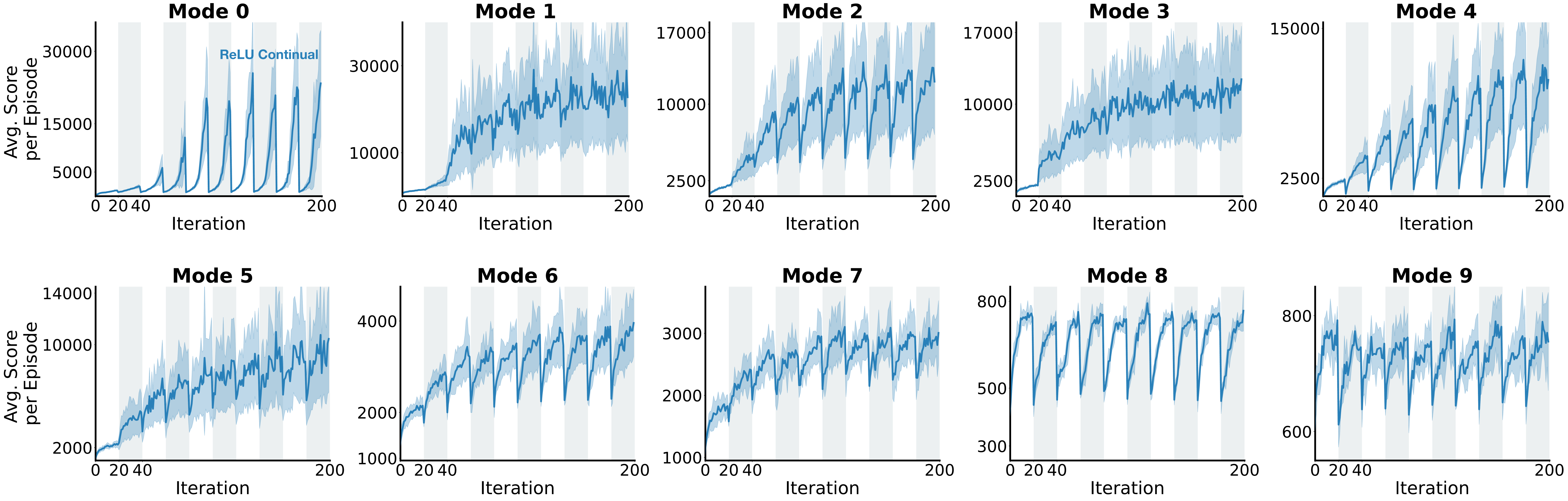}}
\caption{\textbf{Loss of Plasticity was not observed in Space Invaders on a sequence of 10 game modes.} The structure of this plot is identical to Figure~\ref{fig:breakout_10_modes}. The blue curves show Rainbow learning as it cycles through 10 different game modes of Space Invaders in a fixed sequence. The agent interacts with each game mode for 20M frames in every iteration of the cycle; the cycle restarts every 200M frames.}
\label{fig:space_invaders_10_modes}
\end{center}
\end{figure*}

\begin{figure*}[htb!]
\begin{center}
\centerline{\includegraphics[width=\textwidth]{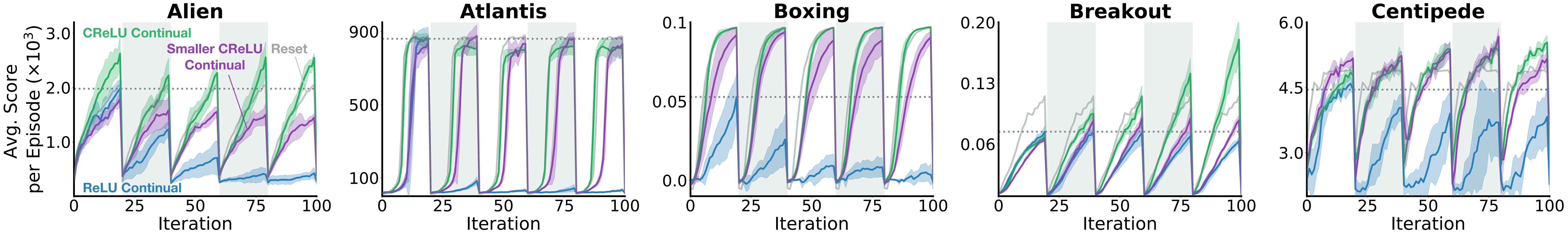}}
\caption{\textbf{Comparison of the performance induced by CReLUs with networks of different sizes.} As discussed in the main text, most plots in the paper were obtained with an invariant input dimension CReLU agent. For this CReLU variant the agent ends up with more parameters than the ReLU agent. Nevertheless, this difference is not the reason for CReLUs success. The purple curve depicts the performance of a Rainbow agent that also use CReLUs as activation functions, but the invariant output dimension one---consequently, this agent with CReLU activations has way fewer parameters than the Rainbow agent with ReLU activations, thus we termed it Smaler CReLU Continual. We can see that although the performance depicted in purple is worse than the one depicted in green, the purple curve, despite being generated by a neural network with way fewer parameters, still does not exhibit loss of plasticity.}
\label{fig:crelus_size}
\end{center}
\end{figure*}

\begin{figure*}
\begin{center}
\centerline{\includegraphics[width=\textwidth]{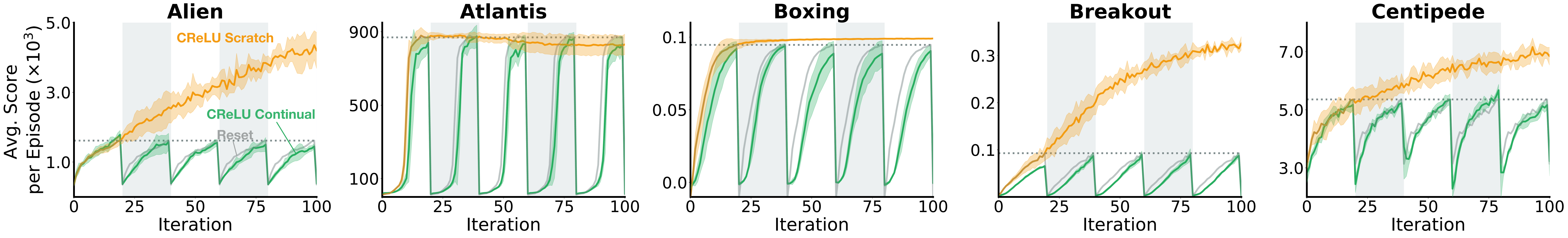}}
\caption{\textbf{CReLUs do not face loss of plasticity.} The orange curve depicts the learning progress of the Rainbow agent when learning from scratch in each game using CReLUs as activation functions, instead of ReLUs. We see the Rainbow agent cycling over multiple games consistently gets back to the final performance the Rainbow agent with CReLU activations reach at the end of the first iteration. Nevertheless, it is clear that the performance of the agent that cycles through multiple games is nowhere near the performance of an agent that learns from scratch in a single game. This plot also adds evidence that CReLUs performance is not due to a change in the neural network architecture as both agents in this plot have the same architecture.}
\label{fig:continual_learning_only_with_crelus_5_games}
\end{center}
\end{figure*}

\begin{figure*}[htb!]
\begin{center}
\centerline{\includegraphics[width=\textwidth]{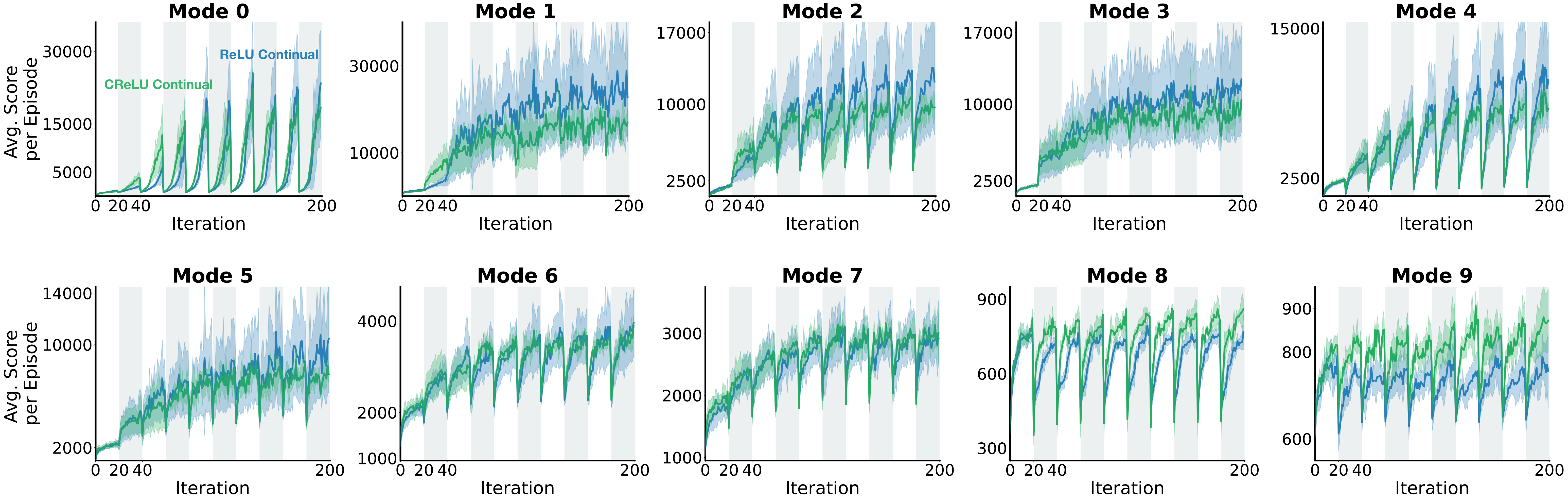}}
\caption{\textbf{Loss of plasticity was not observed in Space Invaders on a sequence of 10 game modes.} The structure of this plot is identical to Figure~\ref{fig:space_invaders_10_modes}. The blue curves show the performance of the Rainbow agent with ReLU activations in the
continual learning setting as it cycles through 10 different game modes of Space Invaders in a fixed sequence. The agent
interacts with each game mode for 20M frames in every iteration of the cycle; the cycle restarts every 200M frames.
The green curves show the performance of Rainbow with CReLU activations in the same continual learning setting. Neither agent exhibits loss of plasticity.}
\label{fig:space_invaders_10_modes_crelu}
\end{center}
\end{figure*}

\begin{figure*}[ht]
\begin{center}
\centerline{\includegraphics[width=\textwidth]{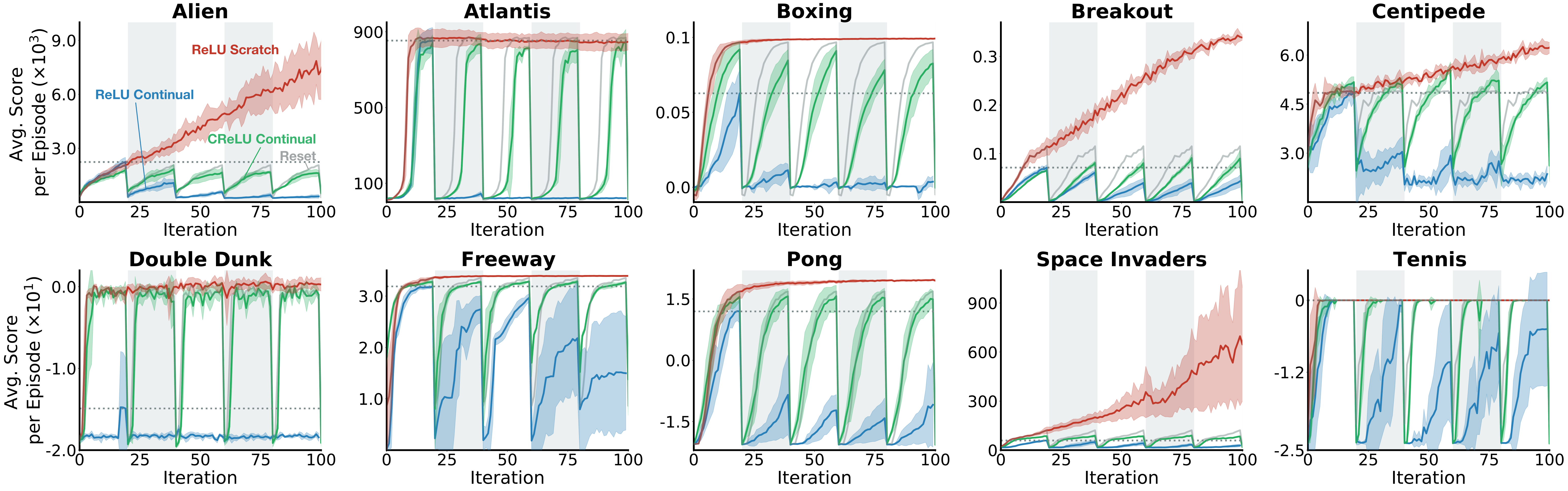}}
\caption{\textbf{Catastrophic forgetting alongside loss of plasticity and CReLU's performance.} This figure reports the same data as Figure~\ref{fig:loss_plasticity_10_games}, with the addition of the learning progress of the Rainbow agent when using CReLUs as activation function (in green). The red curve can be seen as an approximation to the idealized performance of an agent that does not suffer from interference (nor benefits from generalization) from other modes and that does not forget anything about its past, restarting where it left off in the previous visit. Although an unrealistic and unfair baseline, it highlights how much Rainbow is leaving on the table when it cycles through several games, and how the performance achieved when using CReLUs, despite it often surpassing the level of performance the agent achieved the previous time it visited that game, it is still substantially worse than what one can achieve when learning only in a single game.}
\label{loss_plasticity_10_games_final}
\end{center}
\end{figure*}

\end{document}